\documentclass{ifacconf}
\usepackage{amsmath,amsfonts,amssymb}
\makeatletter
\let\old@ssect\@ssect 
\makeatother
\usepackage{natbib}        
\usepackage[hidelinks]{hyperref}
\makeatletter
\def\@ssect#1#2#3#4#5#6{%
\NR@gettitle{#6}
\old@ssect{#1}{#2}{#3}{#4}{#5}{#6}
}
\makeatother
\usepackage{chngcntr}
\counterwithin*{section}{part}
\usepackage{cleveref}                       
\crefname{figure}{fig.}{figs.}              
\Crefname{figure}{Fig.}{Figs.}              
\usepackage{graphicx}                       
\graphicspath{ {./img/} }                   
\usepackage{xcolor}
\usepackage{siunitx}                        
\usepackage{bm}                             
\newcommand{\transposed}{^\mathrm{T}}       
\usepackage{textcomp}                       
\usepackage[gen]{eurosym}                   
\newcommand{\modelname}[1]{#1}
\newcommand{\PTone}{$\textnormal{PT}_{1}$}
\newcommand{\motor}{m}
\newcommand{\servo}{d}
\newcommand{\bat}{u}
\newcommand{\bigO}[1]{\mathcal{O}(#1)}
\newcommand{\vehName}{{\textmu}Car}
\usepackage{acro}                           
\acsetup{single}
\DeclareAcronym{mpc}{
  short = MPC ,
  long  = model predictive control
}
\DeclareAcronym{CPM}{
  short = CPM ,
  long  = cyber-physical mobility
}
\DeclareAcronym{DDS}{
  short = DDS ,
  long  = Data Distribution Service
}
\DeclareAcronym{IPS}{
  short = IPS ,
  long  = indoor positioning system
}
\DeclareAcronym{LLC}{
  short = LLC ,
  long  = low-level controller
}
\DeclareAcronym{MLC}{
  short = MLC ,
  long  = mid-level controller
}
\DeclareAcronym{HLC}{
  short = HLC ,
  long  = high-level controller
}
\DeclareAcronym{NTP}{
  short = NTP ,
  long  = Network Time Protocol
}
\DeclareAcronym{MCU}{
  short = MCU ,
  long  = microcontroller unit
}
\DeclareAcronym{IMU}{
  short = IMU ,
  long  = inertial measurement unit
}
\DeclareAcronym{PCB}{
  short = PCB ,
  long  = printed circuit board
}
\DeclareAcronym{PWM}{
  short = PWM ,
  long  = pulse width modulation
}
\DeclareAcronym{I2C}{
  short = I\textsuperscript{2}C ,
  long  = Inter-Integrated Circuit
}
\DeclareAcronym{SPI}{
  short = SPI ,
  long  = Serial Peripheral Interface
}
\DeclareAcronym{LiPo}{
  short = LiPo ,
  long  = lithium-ion polymer
}
\DeclareAcronym{LDO}{
  short = LDO ,
  long  = low-dropout
}
\DeclareAcronym{GNSS}{
  short = GNSS ,
  long  = Global Navigation Satellite System
}
\DeclareAcronym{BARC}{
  short = BARC ,
  long  = Berkeley Autonomous Race Car
}

\begin{document}
\begin{frontmatter}
\title{Networked and Autonomous Model-scale Vehicles for Experiments in Research and Education\thanksref{footnoteinfo}} 

\thanks[footnoteinfo]{This research is supported by the Deutsche Forschungsgemeinschaft (German Research Foundation) within the Priority Program SPP 1835 ``Cooperative Interacting Automobiles" (grant number: KO 1430/17-1) and the Post Graduate Program GRK 1856 ``Integrated Energy Supply Modules for Roadbound E-Mobility". 
\\\textcopyright\ 2020 the authors. This work has been accepted to IFAC for publication under a Creative Commons Licence CC-BY-NC-ND.
}


\author[First]{Patrick Scheffe}
\author[First]{Janis Maczijewski}
\author[First]{Maximilian Kloock}
\author[First]{Alexandru Kampmann}
\author[First]{Andreas Derks}
\author[First]{Stefan Kowalewski}
\author[First]{Bassam Alrifaee}
\address[First]{Chair for Embedded Software, RWTH Aachen University, 52074~Aachen, Germany (corresponding e-mail: scheffe@embedded.rwth-aachen.de)}

\begin{abstract}                
This paper presents the \vehName, a 1:18 model-scale vehicle with Ackermann steering geometry developed for experiments in networked and autonomous driving in research and education. The vehicle is open source, moderately costed and highly flexible, which allows for many applications. It is equipped with an \acl{IMU} and an odometer and obtains its pose via WLAN from an \acl{IPS}. 
The two supported operating modes for controlling the vehicle are (1) computing control inputs on external hardware, transmitting them via WLAN and applying received inputs to the actuators and (2) transmitting a reference trajectory via WLAN, which is then followed by a controller running on the onboard Raspberry Pi Zero W.
The design allows identical vehicles to be used at the same time in order to conduct experiments with a large amount of networked agents.
\end{abstract}


\end{frontmatter}
\section*{Supplementary material}
A demonstration video of this work is available at \url{https://youtu.be/aH1Q8AKXmUs}.

The vehicle software, bill of materials and a production tutorial is referenced from our website \url{http://cpm.embedded.rwth-aachen.de}.

\section{Introduction}
\label{sec:intro}

Research on networked and autonomous vehicles is ongoing since multiple decades. When new methods are developed, the necessity of testing them arises. This can be done with little effort in simulation as in \cite{naumann2018}. The meaningfulness of results in simulation is restricted, as only aspects of reality that are modeled are considered. More meaningful are experiments in true scale, but those require a high effort and are expensive, especially when testing methods on networked vehicles, as multiple test platforms are required. Midway between those options, methods can be tested on scaled testbeds. In scaled experiments, many challenges of the true-scale problem are apparent, e.g. communication delays and losses, synchronization problems or actuator dynamics. Another benefit compared to the true-scale experiment is that setting up the experiment is simpler and quicker, which allows for rapid development cycles. 

The curriculum at a university should prepare students for research in networked and autonomous vehicles. This includes for example the design of algorithms for embedded hardware, designing controllers for nonlinear systems, or coupling of networked agents for collision avoidance. Seeing an algorithm one has developed running in an experiment fills students with enthusiasm about learning concepts of control by applying it to the \ac{CPM} system. The modified model-scale vehicle proposed in this paper enables those experiments.

This paper is structured as follows.\Cref{sec:existing_platforms} compares model-scale vehicles with Ackermann steering geometry from literature. \Cref{sec:hw} describes how we transform a model-scale race car to a networked and autonomous vehicle with off-the-shelf components, excluding a printed circuit board. The lab environment in which the vehicles operate is sketched in \cref{sec:env}. In \cref{sec:vce}, examples are given to show in what form the vehicles can be used in control education.

\section{Existing platforms}
\label{sec:existing_platforms}
In the last decade, a number of model-scale testbeds have been developed. In \cite{paull2017}, 15 platforms for education and research with a cost lower than \SI{300}[\$]{} are compared. These differ from the model-scale vehicle we present, as they are wheeled differential drive platforms or platforms with slip-stick forwards motion.

In \cref{tab:rel_veh}, an overview of recently developed model-scale vehicles with Ackermann steering geometry is given. Having a scaling factor of 1:43 and 1:24 respectively, the ORCA Racer \cite{liniger2014} and the Cambridge Minicar \cite{hyldmar2019} are smaller than the vehicle presented in this work.
The ORCA Racer is based on the Kzosho \modelname{dnano} RC race car, but substitutes its original board with a custom \ac{PCB}. This board features an ARM Cortex-M4 processor, Bluetooth communication and an \ac{IMU}. The vehicles are designed to receive externally computed control inputs via Bluetooth, and apply these inputs with an onboard \ac{LLC}.
The Cambridge Minicar is based on the CMJ RC Cars \modelname{Range Rover Sport}. Its controlled by a Raspberry Pi Zero W. These vehicles are controlled by sending externally computed control inputs via broadband radio.\\
The \ac{BARC} from \cite{gonzales2016}, the MIT Racecar from \cite{karaman2017} and the F1/10 from \cite{okelly2019} share the scale of 1:10.
The mechanical base for all three vehicles is a Traxxas rally car. At this size, the vehicles are capable of carrying more computational power and more sensors additionally to an \ac{IMU}.
In the \ac{BARC}, 4 rotary encoders are installed for speed measurement and a camera is mounted. Optionally, it is possible to install a lidar and a \ac{GNSS} receiver. The \ac{HLC} and main computing unit is an ODROID-XU4, the \ac{LLC}, i.e. sensor read and actuator control, is performed with an Arduino Nano.
The setup of the MIT Racecar and the F1/10 is similar. The speed is given by a VESC electronic speed controller, and optional sensors include a 3D stereo cameras and a lidar. The main computing element is the Nvidia Jetson Tegra X1.
The greater computing power and additional sensors allow for onboard autonomy. This is also a reason why these setups cost around \SI{1000}[\$]{}. At the scale of 1:10, a lot of space is required for indoor experiments on cooperative driving with multiple vehicles. Due to the cost and the size of the platforms, indoor experiments with a large amount of vehicles are difficult.\\
At the largest scale of 1:5, the GATech Auto-Rally from \cite{williams2016} and the IRT buggy from \cite{reiter2014} and \cite{reiter2017} are designed for outdoor experiments.
The Auto-Rally is equipped with two forward facing cameras, a Lord Microstrain 3DM-GX4-25 \ac{IMU}, a \ac{GNSS} receiver, and wheel speed sensors. The computational power is provided by an Intel quad-core i7 processor, 16GB RAM, and an Nvidia GTX-750ti graphics card. With this elaborate hardware setup, the Auto-Rally is used for aggressive driving.
The IRT buggy is designed for versatile use. It shares the separation of \ac{HLC} and \ac{LLC} in two hardware components with the \ac{BARC}. Sensors include a \ac{GNSS}-sensor, an \ac{IMU}, and two rotary encoders at the rear wheels. Its modular setup allows for other sensors such as a lidar. Similar to the ORCA Racer, this platform is not open source.
\begin{table}
    \caption{Recent model-scale Ackermann-steering platforms}
    \centering
    \begin{tabular}{l | l }
        Vehicle name       & Scale \\
        \hline
        ETHZ ORCA Racer     & 1:43 \\
        Cambridge Minicar   & 1:24 \\
        \emph{\vehName}     & \emph{1:18}\\
        F1/10               & 1:10 \\
        \ac{BARC}           & 1:10 \\
        MIT Racecar         & 1:10 \\
        GATech AutoRally    & 1:5  \\
        IRT buggy           & 1:5  \\
    \end{tabular}
    \label{tab:rel_veh}
\end{table}

The larger model-scale vehicles are equipped with sensors and computing power to allow autonomy. The \vehName, as well as the ORCA Racer and the Cambridge Minicar are reliant on the interaction with a lab environment. This lab environment provides the positioning of the vehicles and therefore substitutes the \ac{GNSS} of the real world experiment. In the case of the Cambridge Minicar, this is done with an OptiTrack motion capture system that requires multiple cameras, while the lab environment of the ORCA Racer only uses one camera, similar as our \ac{CPM} lab. In contrast to those two labs, in addition to the option of sending control inputs to the vehicle, a trajectory following mode exists, where an onboard controller determines the control inputs necessary to follow a given trajectory.

\section{Vehicle setup} 
\label{sec:hw}
\begin{figure}
    \centering
    \includegraphics[width=\linewidth]{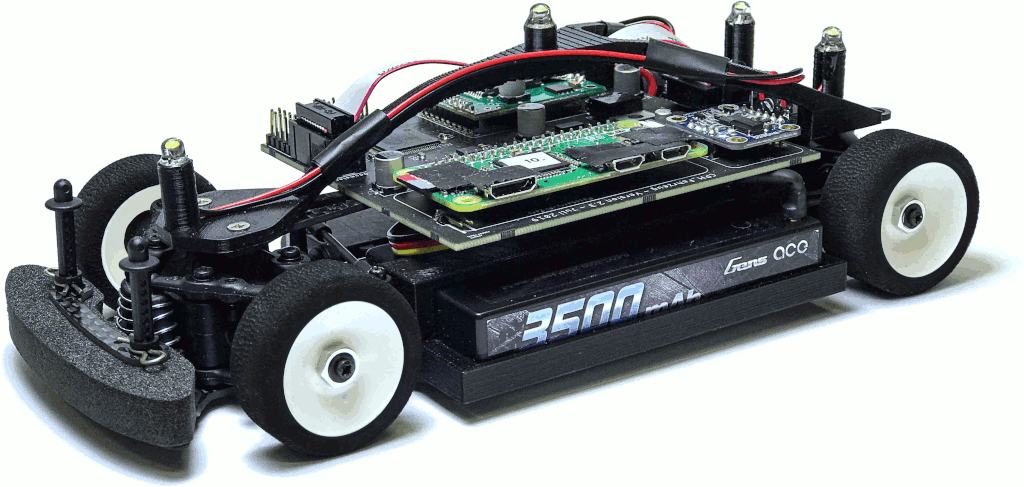}
    \caption{The \vehName, a 1:18 model-scale vehicle}
    \label{fig:veh_iso}
\end{figure}
The model-scale vehicle presented here is shown in \cref{fig:veh_iso}. It is an Ackermann-steered, non-holonomic mobile robot in the scale of 1:18 compared to a typical passenger vehicle. Its length is \SI{220}{mm}, its width \SI{107}{mm}, its height \SI{70}{mm}, its wheelbase $ L = \SI{150}{mm}$ and its weight is \SI{500}{g}. The vehicle has a maximum speed of \SI{3.7}{m \per s}. The power consumption in standby (without steering or acceleration) is \SI{250}{mW}. In experiments, the battery powers the car for about five hours. \cref{tab:veh_cost} lists the components used in the model-scale vehicle. The cost calculation refers to an order of 20 vehicles, as a single \ac{PCB} would cost \SI{45}{\euro}, but ordering a panel cluster with 20 \acp{PCB} on one board reduces the price of a unit to \SI{15}{\euro}. Assembling a vehicle takes one person around six hours of time.
\begin{table}
    \caption{Components used in the \vehName; cost rounded to the next integer}
    \centering
    \begin{tabular}{l | l | S}
        Item                    & Application       & {Cost [\euro]}\\
        \hline
        XRAY M18 Pro            & Mechanical platform   & 170 \\
        Gens ace 3500mAh        & LiPo Battery          & 30 \\
        NF113LG-011             & Motor                 & 15 \\
        Hitec D89MW             & Servo                 & 50 \\
        PCB	                    & Board                 & 15    \\
        Raspberry Pi Zero W     & MLC                   & 18 \\
        8GB SD Card             & Memory                & 7  \\
        ATmega2560              & LLC                   & 12 \\
        Pololu VNH5019          & Motor Driver          & 23 \\
        DeboSens BNO055         & IMU                   & 34 \\
        Eletronic Parts         &                       & 21 \\
        \hline                                          
        SUM                     &                       & 395 \\
    \end{tabular}
    \label{tab:veh_cost}
\end{table}

Using an off-the-shelf mechanical platform allows for a quick start in building a networked and autonomous model-scale vehicle. We use the mechanical components from the XRAY \modelname{M18 PRO LiPo}. It is a 1:18 micro car that is designed for holding a battery, a servo motor for steering and a motor for propulsion. The motor drives all four wheels as the shaft is connected to each one with differentials. The minimum turning radius given by the mechanical design is approximately \SI{0.3}{m}.

\begin{figure}
    \centering
    \resizebox{\linewidth}{!}{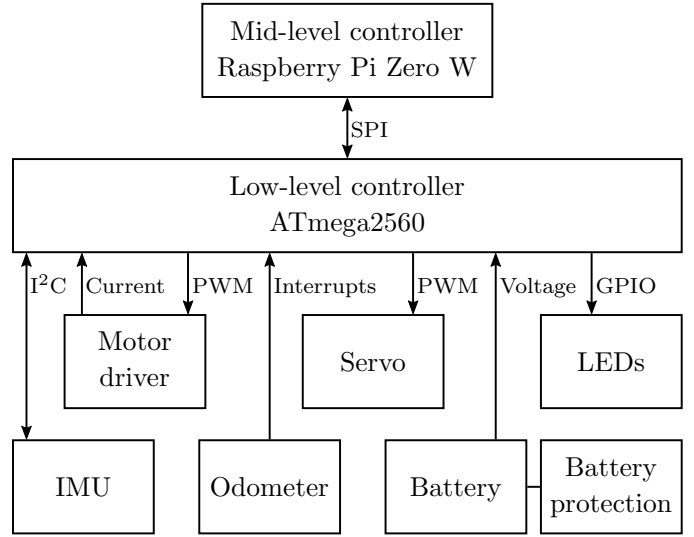}
    \caption{Vehicle hardware architecture}
    \label{fig:veh_hw}
\end{figure}

\begin{figure}
    \centering
\begingroup%
  \makeatletter%
  \providecommand\color[2][]{%
    \errmessage{(Inkscape) Color is used for the text in Inkscape, but the package 'color.sty' is not loaded}%
    \renewcommand\color[2][]{}%
  }%
  \providecommand\transparent[1]{%
    \errmessage{(Inkscape) Transparency is used (non-zero) for the text in Inkscape, but the package 'transparent.sty' is not loaded}%
    \renewcommand\transparent[1]{}%
  }%
  \providecommand\rotatebox[2]{#2}%
  \newcommand*\fsize{\dimexpr\f@size pt\relax}%
  \newcommand*\lineheight[1]{\fontsize{\fsize}{#1\fsize}\selectfont}%
  \ifx\svgwidth\undefined%
    \setlength{\unitlength}{245.71811706bp}%
    \ifx\svgscale\undefined%
      \relax%
    \else%
      \setlength{\unitlength}{\unitlength * \real{\svgscale}}%
    \fi%
  \else%
    \setlength{\unitlength}{\svgwidth}%
  \fi%
  \global\let\svgwidth\undefined%
  \global\let\svgscale\undefined%
  \makeatother%
  \begin{picture}(1,0.60943733)%
    \lineheight{1}%
    \setlength\tabcolsep{0pt}%
    \put(0,0){\includegraphics[width=\unitlength,page=1]{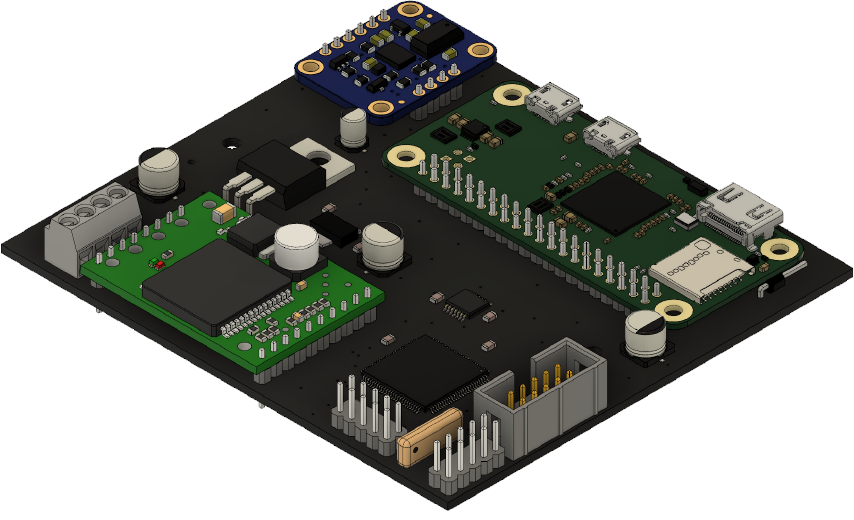}}%
    \put(0.31731435,0.56996834){\color[rgb]{0,0,0}\makebox(0,0)[rt]{\lineheight{1.25}\smash{\begin{tabular}[t]{r}IMU\end{tabular}}}}%
    \put(0.15672738,0.00840569){\color[rgb]{0,0,0}\makebox(0,0)[lt]{\lineheight{0.29999998}\smash{\begin{tabular}[t]{l}ATmega2560\end{tabular}}}}%
    \put(0.69822602,0.56222839){\color[rgb]{0,0,0}\makebox(0,0)[lt]{\smash{\begin{tabular}[t]{l}Raspberry Pi\\Zero W\end{tabular}}}}%
    \put(0,0){\includegraphics[width=\unitlength,page=2]{printed_circuit_board.pdf}}%
    \put(0.03021397,0.15126118){\color[rgb]{0,0,0}\makebox(0,0)[lt]{\smash{\begin{tabular}[t]{l}Motor\\driver\end{tabular}}}}%
    \put(0.66546897,0.04443122){\color[rgb]{0,0,0}\makebox(0,0)[lt]{\smash{\begin{tabular}[t]{l}Odometer board\\connector slot\end{tabular}}}}%
  \end{picture}%
\endgroup%

    \caption{The \ac{PCB} on the vehicle with several components installed}
    \label{fig:veh_pcb}
\end{figure}

The vehicle's hardware architecture is illustrated in \cref{fig:veh_hw}.
A Raspberry Pi Zero W takes the role of the \ac{MLC} on the vehicle. It is responsible for the communication via WLAN with the \ac{HLC}, as described in \cref{sec:env}, and for clock synchronization using the \ac{NTP}. Additionally, the \ac{MLC} fuses the sensor data to obtain accurate localization. The \ac{MLC} also supplies the \ac{LLC} with control inputs. This is either realized by forwarding control inputs received via WLAN, or by running a controller for trajectory following as described in the next paragraph. The tasks on the Raspberry are repeated in a frequency of \SI{50}{Hz}, i.e. a time interval of \SI{20}{ms}.\\
In order to ensure the most individual and adaptable handling of the vehicle, we designed a custom \ac{PCB} connecting the components.. This \ac{PCB} serves as an interface between the actuators, sensors and control electronics. The \ac{PCB} with its components is shown in \cref{fig:veh_pcb}. 
This \ac{PCB} embeds an \modelname{ATmega2560} microcontroller with a \SI{16}{MHz} clock rate. This microcontroller represents the \ac{LLC}, reading the sensor data and applying the control inputs to the actuators. The hardware separation in \ac{MLC} and \ac{LLC} introduces a hierarchical architecture, which creates a hardware abstraction layer. Even in the case the \ac{MLC} is changed, the interface to the hardware will stay the same.\\
In a frequency of \SI{50}{Hz}, the \ac{MLC} and the \ac{LLC} exchange information via \ac{SPI}. The \ac{MLC} provides the control inputs, while the \ac{LLC} returns the sensor readings. A TXB0104 bidirectional voltage-level translator was installed for level adaptation of the \ac{SPI} bus. The 3.3V \ac{SPI} level of the Raspberry is converted into a 5V \ac{SPI} signal for the ATmega.\\
The \ac{IMU} is a \modelname{DeboSens BNO055} and provides the required sensor data using a 9-DOF sensor. The ATmega microcontroller can retrieve this data via the two wire \ac{I2C} bus.\\
The motor driver board \modelname{VNH5019} drives the single brushed DC motor of the vehicle via an integrated H\nobreakdash-Bridge. The ATmega controls the engine driver via a \ac{PWM} signal with a frequency of up to \SI{20}{kHz}. A current sensing output provides the ATmega with a signal which is proportional to the current applied to the motor.
The power source is a \SI{2000}{mAh} \ac{LiPo} battery which provides a \SI{7.4}{V} voltage. This voltage is directly fed to the motor driver unit. Since the Raspberry and all the other components (except the motor driver unit) are specified to \SI{5}{V} or \SI{3.3}{V} respectively, the voltage is reduced by an NCP1117 \ac{LDO} voltage regulator. To protect the \ac{LiPo} battery as well as the electronic components a battery protection circuit was inserted.\\
Three Hall-effect sensors mounted on a separate odometer board measure the motor shaft rotation. A diametrically polarized magnet is attached to the motor shaft in order to make the rotational motion of the axis electrically visible. With this setup, it is possible to distinguish six different motor angles per rotation. The digital signals of the Hall sensors are directly transmitted to three I/Os of the ATmega, which translates the signals into rotation ticks.\\
Four LEDs are installed on the vehicle, which are also connected to the odometer board and controlled by the ATmega. The outer three LEDs are used for positioning with an \ac{IPS}, while the inner one communicates the vehicle's ID.

The vehicles can operate in the two different modes (1)~external control and (2)~trajectory following.
If a trajectory is provided to the vehicle, the \ac{MLC} determines the control inputs to follow that trajectory. The trajectory is provided as a list of tuples $ (t, x, y, v_x, v_y) $. Usually, a trajectory point is understood to be a tuple of time and position. The controller needs reference trajectory points at controller-specific points in time. If the time step between trajectory points is assumed to be larger than the control time step, the \ac{MLC} interpolates the trajectory to determine a sensible reference point. By fixing the derivative of the trajectory in each point, the \ac{MLC} is enabled to interpolate between trajectory points with cubic Hermite splines. Additionally, if new reference trajectory points are transmitted, the \ac{MLC}-internal reference trajectory will not change.
If the \ac{MLC} receives control inputs, it switches to directly applying those to the actuators.
This behavior allows for manual control of the vehicle with a gamepad or a keyboard for example.
It is also possible to compute control inputs depending on the vehicle state and reference trajectory externally and send those via WLAN.

\section{Environment: \ac{CPM} lab}
\label{sec:env}
\begin{figure}
    \centering
    \includegraphics[width=\linewidth]{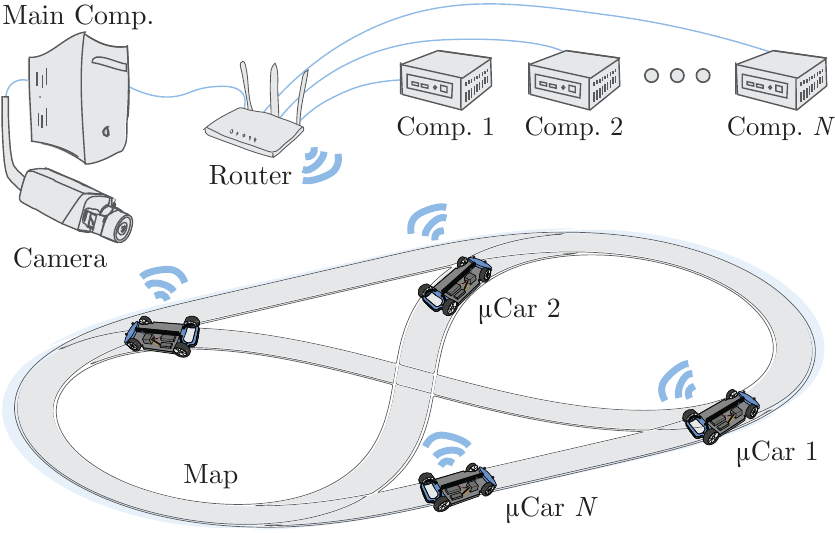}
    \caption{\ac{CPM} lab overview: vehicles communicate via WLAN with their respective computers and the \ac{IPS}.}
    \label{fig:lab_overview}
\end{figure}
As mentioned earlier, the vehicles are used for experiments in a lab environment as visualized in \cref{fig:lab_overview}, which we call \ac{CPM} lab. This lab provides a driving area of about $ 4.5\si{m}\times 4 \si{m}$.
Communication between the vehicles and this environment is established through \ac{DDS} \modelname{RTI Connext DDS}.
An \ac{IPS} provides the vehicles with their pose (position and orientation) with a worst-case accuracy of \SI{3.25}{cm} and \SI{2.25}{\degree}. A camera detects the position of the three LEDs on the vehicle. These LEDs define a vehicle's pose due to their arrangement on the vehicle in a non-equilateral triangle. The vehicle corresponding to a detected pose is identified with a signal code sent by the fourth LED on the vehicle as shown in \cite{kloock2020}. Additionally, a reference trajectory or the actuator inputs for the vehicles are sent via WLAN. The vehicle returns its current state, which includes the estimated pose as well as sensor readings and actuator commands.

\section{The vehicles in control education}
\label{sec:vce}
The vehicle's hierarchical architecture allows students to work at different levels of abstraction.
\begin{enumerate}
    \item It is possible to learn the basics of embedded programming when working with the \ac{LLC} (the \modelname{ATmega2560}). At this level, students need to understand \ac{MCU} data sheets in order to determine how to read sensors and control actuators correctly in C-code.
    \item At the level of the \ac{MLC} (the \modelname{Raspberry Pi Zero W}), tasks like trajectory control or sensor fusion can be tackled.
    Measurements of multiple sensors need to be fused for vehicle localization in the proposed setup, which reflects the real world application. The \ac{IPS} provides absolute positioning, but its measurement data is transmitted to the vehicle via WLAN, which makes the measurements relatively slow and also unreliable. On the other hand, onboard sensors like the \ac{IMU} and the odometer are fast and accurate for short distances, but need a reference.
    A controller for trajectory following can be implemented as simple as a PID-controller, or more advanced as a \ac{mpc}. The \vehName currently uses \ac{mpc} for trajectory following.
    Restrictions by the limited computation power of the \ac{MLC} still apply, which motivates efficient algorithms and a programming language like C\texttt{++}.
    \item On the highest abstraction level, ideas can be developed on an external PC with programming languages common in optimization (e.g. MATLAB, Python). It is possible to work on trajectory planners as well as on external controllers for the vehicles, depending on which mode of operation one wishes to use.
\end{enumerate}

The modularity allows to focus on one specific part of networked and autonomous vehicles. It is possible to provide necessary interfaces with working components, so the content to be taught can be chosen freely and appropriately.

A basis for many control tasks is an appropriate model of the system. A system model is useful for e.g. simulation or controller design. The purpose of the model defines its requirements. For simulation, the goal might be to represent the system as truthfully as possible, while for a controller using \ac{mpc} the ability for fast computation might be necessary.
Since having a system model is the prerequisite of many aspects in control, we show an example of how a model for the model-scale vehicles can be obtained. The goal of this endeavor is to illustrate how the vehicles might serve as a platform to control engineering education.

\subsection{Vehicle dynamics model}
In this example, we aim for a model that is suitable for \ac{mpc} of a vehicle's pose and velocity on embedded hardware. The model needs to be simple enough for quick computation, while accurate enough for predicting the states. We propose a kinematic bicycle model with some added terms to account for various errors.


The model has the states $ \bm{x} $ and inputs $ \bm{u} $
\begin{equation}
    \begin{aligned}
        \bm{x} &= ( x\quad y\quad \psi\quad v)\transposed \\
        \bm{u} &= ( \motor\quad \servo\quad \bat)\transposed,
    \end{aligned}
\end{equation}
where $ x $ and $ y $ are the x- and y-position respectively, $ \psi $ is the yaw angle, $ v $ the speed at the vehicle rear axle, $ \motor $ the dimensionless motor command, $ \servo $ the dimensionless steering command and $ \bat $ the battery voltage. The battery voltage is of course not an input set by the controller, but one that affects the system dynamics.
\begin{figure}
    \begin{center}
        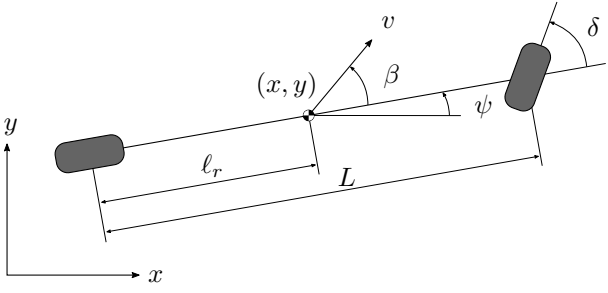
        \caption{Kinematic bicycle model of the vehicle} 
        \label{fig:bicycle_model}
    \end{center}
\end{figure}
The model used to describe the vehicle's dynamics is a non-linear kinematic bicycle model according to \cite{rajamani2011}. Similar to \cite{alrifaee2017}, it is assumed that no slip occurs on the front and rear wheels, and no forces act on the vehicle. The velocity dynamics are described with a \PTone\ behavior, which results in the following equations
\begin{equation}
\begin{aligned}
    \dot{x}     ={} & v \cdot \sqrt{1+\left( \dfrac{\ell_r}{L}\cdot \tan\delta \right)^2} \cdot \cos(\psi + \beta) \\
    \dot{y}     ={} & v \cdot \sqrt{1+\left( \dfrac{\ell_r}{L}\cdot \tan\delta \right)^2} \cdot \sin(\psi + \beta) \\
    \dot{\psi}  ={} & v \cdot \frac{1}{L} \cdot \tan\delta \\
    \dot{v}     ={} & -\dfrac{1}{T_{v}} \cdot v + \dfrac{K_{v}}{T_{v}} \cdot v_{\textnormal{in}}(\motor, \bat) \\
    \beta       ={} & \tan^{-1}\left(\frac{\ell_r}{L} \tan\delta \right).
\end{aligned}
\end{equation}
The model variables are illustrated in \cref{fig:bicycle_model}. $ \ell_r $ is the distance from the rear axle to the vehicle's reference point, $ L $ is the distance between front and rear axle, $\delta$ is the steering angle which is related to the steering command $d$, $ K_{v} $ and $ T_{v} $ are the gain and time constant of the velocity's \PTone\ behavior and $ v_{\textnormal{in}} $ is the input velocity, which is modelled as a function of the motor command $ \motor $ and the battery voltage $ u $.
The change of the vehicle's $ x $- and $ y $-position is dependent on the velocity $ v_{c} $ at the vehicle's reference point. From the fact that the angular velocity $ \dot{\psi} $ is equal at every point of the vehicle, we get
\begin{equation}\label{eqn:dotpsi_const}
    \dot{\psi} =  \dfrac{v_{c}}{R_c} = \dfrac{v}{R},
\end{equation}
where $ R_{c} $ and $ R $ are the radii of the circular movement at the vehicle center and rear axle respectively. With \eqref{eqn:dotpsi_const} and Pythagoras' theorem we obtain
\begin{equation}\label{eqn:vc_v}
    v_c = v \cdot \sqrt{1+\left( \dfrac{\ell_r}{L}\cdot \tan\delta \right)^2}.
\end{equation}

In order to simplify computational tasks on the model, we can approximate some terms with Taylor series at the point $ \delta = 0 $.
The side slip angle $ \beta $ due to steering is approximated with a first-order Taylor series
\begin{equation}\label{eqn:side_slip_taylor}
\beta (\delta) = \dfrac{\ell_r}{L} \cdot \delta + \bigO{\delta^3}.
\end{equation}
Equation~\eqref{eqn:vc_v} is simplified with a second-order Taylor approximation:
\begin{equation}\label{eqn:vc_v_taylor}
v_c = v \cdot \left( 1 + \left(\dfrac{\ell_r}{L}\right)^2 \cdot \delta^2 + \bigO{\delta^4}\right).
\end{equation}
Now substituting the model's variables with parameters and introducing some parameters to account for various inaccuracies, the parameterized bicycle model is given by:
\begin{equation}\label{eqn:model_simplified}
\begin{aligned}
\dot{x}   ={} & p_1 \cdot v \cdot \left(1+p_2 \cdot (\servo + p_{9})^2\right) \\
                & \cdot \cos(\psi + p_3 \cdot (\servo + p_{9}) + p_{10}) \\
\dot{y}   ={} & p_1 \cdot v \cdot \left(1+p_2 \cdot (\servo + p_{9})^2\right) \\
                & \cdot \sin(\psi + p_3 \cdot (\servo + p_{9}) + p_{10}) \\
\dot{\psi}  ={} & p_4 \cdot v \cdot (\servo + p_{9}) \\
\dot{v}     ={} & p_5 \cdot v + (p_6 + p_7 \cdot \bat) \cdot \text{sign}(\motor) \cdot |\motor|^{p_8}.
\end{aligned}
\end{equation}
An extra parameters introduced is $p_1$, which compensates the calibration error between \ac{IPS} speed and odometer speed. $ p_{2} $ and $ p_3 $ substitute the model parameters in \eqref{eqn:vc_v_taylor} and \eqref{eqn:side_slip_taylor} respectively. $ p_{4} $ takes care of the model parameter $ \frac{1}{L} $ as well as the conversion of steering command to steering angle.
$ p_5 $ substitutes $ -\frac{1}{T_{v}} $in the velocity's \PTone\ model. The steady state velocity is modeled as a power function, where the constant factor is represented by $ p_{6} $ and the exponent by $ p_{8} $. In order to avoid the trouble that comes with negative bases and real exponents, the absolute value of the motor command $ \motor $ is used as the base and the sign of $ \motor $ is multiplied. As the motor strength depends on the battery voltage, we added it as a multiplying factor with the parameter $ p_{7} $.
$p_{9}$ is an extra parameter introduced to correct steering misalignment, while $p_{10}$ accounts for a yaw calibration error in the \ac{IPS}.

This is an end-to-end, grey-box model for the vehicle dynamics. The model parameters are not measured directly, but optimized to best fit the vehicle behavior as shown in \cref{sec:vce:pi}.

\subsection{Model discretization}
The model is discretized with the explicit Euler method, as follows:
\begin{equation} \label{eqn:model_discretized}
    \bm{x}_{k+1} = \bm{x}_k + \Delta t \cdot f(\bm{x}_k,  \bm{u}_k, \bm{p}).
\end{equation}
Here, $f$ is obtained from the continuous vehicle dynamics model \eqref{eqn:model_simplified}. This discretization is chosen for its simplicity and computational efficiency. Measurements are taken in time intervals of $\Delta t = \SI{0.02}{s}$. This short time interval compensates partly for the inaccuracies introduced by the method used, and the discretization is included in the parameter identification.

\subsection{Parameter identification}
\label{sec:vce:pi}
Since the dynamics of nonholonomic vehicles are nonlinear, model identification procedures for nonlinear systems need to be used.
Identifying the vehicle dynamics can be achieved by formulating the task as an optimal parameter estimation problem. The optimization tries to find a set of model parameters that best reproduce the measurement data. 
A measurement vector at timestep $ k $ contains:
\begin{equation}
    \hat{\bm{x}}_{k} = (\hat{x}_{k}\quad \hat{y}_{k}\quad \hat{\psi}_{k}\quad \hat{v}_{k})\transposed.
\end{equation}
Here, $\hat{x}$ and $\hat{y}$ is the \ac{IPS} x- and y-position respectively, $\hat{\psi}$ is the \ac{IPS} yaw angle and $\hat{v}$ the odometer speed.

The optimization problem is then given as
\begin{equation}
\begin{aligned}
    \underset{\bm{x}_k^j, \bm{p}}{\text{minimize}} && 
        \sum_{j=1}^{n_{\text{experiments}}} \sum_{k=0}^{n_{\text{timesteps}}} E(\bm{x}_k^j - \hat{\bm{x}}_k^j) \\
    \text{subject to} && 
        \begin{aligned}[t]
            \bm{x}_{k+1}^j &= \bm{x}_k^j + \Delta t \cdot f(\bm{x}_k^j,  \bm{u}_k^j, \bm{p}) \\
            k &=0,\mathellipsis, (n_{\text{timesteps}}-1) \\
            j &=1,\mathellipsis, n_{\text{experiments}},
        \end{aligned}
\end{aligned} 
\end{equation}
where $\bm{u}_k^j$ are the measured inputs, $f$ is the discrete vehicle model as in \eqref{eqn:model_discretized}, $\bm{p}$ is the vector of model parameters $ p_{1} $ to $ p_{10} $, $\Delta t$ is a constant timestep of \SI{0.02}{s} and $E$ is the error penalty function.
Since the vehicle pose lives in $SE(2)$, an adequate error metric needs to be used. We used a weighted quadratic error function and accounted for the period of $2\pi$ in the yaw error function using $\sin^2(\Delta\psi/2)$.

This kind of optimization problem is not well suited for identifying the delay times. The optimization problem is therefore solved multiple times for combinations of delay times in an outer loop. The delays that create the lowest objective value are taken as the solution.

\begin{figure}
    \begin{center}
        \includegraphics[width=0.8\linewidth]{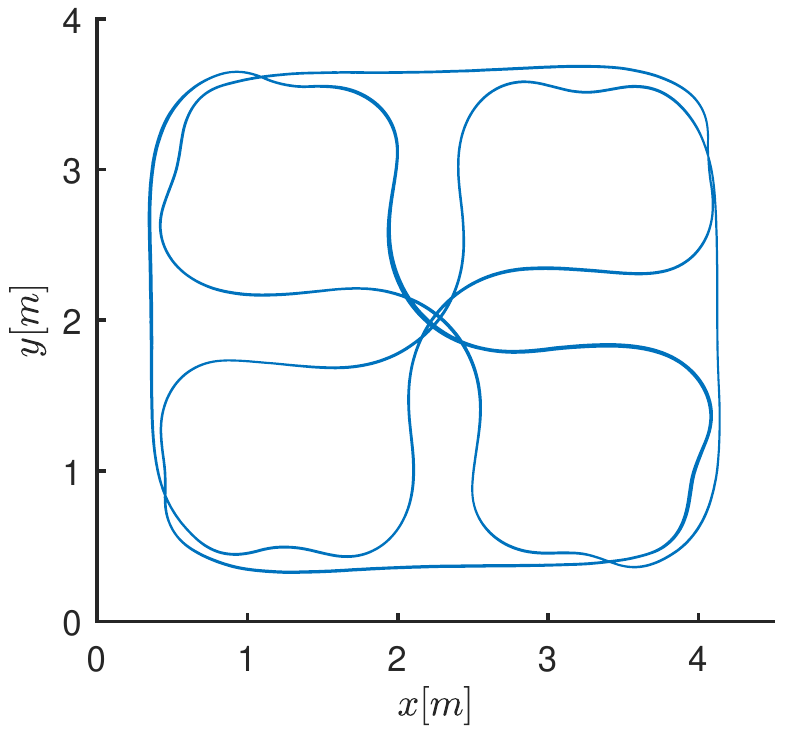}
        \caption{Driven trajectory for measurement data collection} 
        \label{fig:meas1}
    \end{center}
\end{figure}
The measurement data used in the parameter optimization is shown in \cref{fig:meas1}. This data is sliced into parts of 100 consecutive data points, i.e. time intervals of \SI{2}{s}, which are fed to the optimization problem as experiments. The resulting parameters are
\begin{equation}
\begin{aligned}
\bm{p} = 
    ( 
        &&  1.00 && -0.14 && 0.20 && 3.56 && -2.19& &&\\
        && && -9.73 &&  2.52 && 1.32 && 0.03 && -0.01&
    ).
\end{aligned}
\end{equation}
The delays identified are 1 timestep for the IPS data, 0 timesteps for the local measurement information and 5 timesteps for the motor and steering actuation.

\section{Conclusion}
\label{sec:conclusion}
This paper presented how a regular RC race car can be transformed to a networked and autonomous vehicle with mainly off-the-shelf components. The vehicles are used for teaching in multiple courses at RWTH Aachen University at the moment. We are eager to see the impact of applying concepts on real control systems on the students' learning experience. 


Currently, a fleet of 20 vehicles is being built up. This should enable students and researchers alike to perform various experiments on networked and autonomous driving in moderately large scale networked systems. 


\bibliography{ifacconf}     

\pagebreak
\appendix
\section{Required Demonstrator Space}
The 1:18 model-scale vehicles will be presented with a reduced lab environment. For that, we need 
\begin{enumerate}
    \item space of about $ \SI{1}{m}\times \SI{1.5}{m} $ and 
    \item power outlets.
\end{enumerate}

\end{document}